# PERC: a suite of software tools for the curation of cryoEM data with application to simulation, modelling and machine learning

**Beatriz Costa-Gomes[a†], Joel Greer[b†], Nikolai Juraschko[a,c,d†], James Parkhurst[c,e†], Jola Mirecka[b], Marjan Famili[a], Camila Rangel-Smith[a], Oliver Strickson[a], Alan Lowe[a], Mark Basham[c*], Tom Burnley[b*]**

[a] The Alan Turing Institute, British Library, 96 Euston Road, London, England, NW1 2DB, United Kingdom

[b] Science & Technology Facilities Council, Research Complex at Harwell, Oxon, OX11 0FA, United Kingdom

[c] Rosalind Franklin Institute, Harwell Science & Innovation Campus, Didcot, Oxfordshire, OX11 0QX, United Kingdom, d University of Oxford, Oxford, OX1 3QU, United Kingdom

[e] Diamond Light Source, Harwell Science & Innovation Campus, Didcot, Oxfordshire, OX11 0DE, United Kingdom

[†]Authors contributed equally

*Correspondence e-mail: mark.basham@rfi.ac.uk, tom.burnley@stfc.ac.uk

**Synopsis** PERC (profet, EMPIARreader and CAKED) a suite of open source python software tools facilitating the curation of cryoEM data utilising standard data science libraries.

**Abstract** Ease of access to data, tools and models expedites scientific research. In structural biology there are now numerous open repositories of experimental and simulated datasets. Being able to easily access and utilise these is crucial for allowing researchers to make optimal use of their research effort. The tools presented here are useful for collating existing public cryoEM datasets and/or creating new synthetic cryoEM datasets to aid the development of novel data processing and interpretation algorithms. In recent years, structural biology has seen the development of a multitude of machine-learning based algorithms for aiding numerous steps in the processing and reconstruction of experimental datasets and the use of these approaches has become widespread. Developing such techniques in structural biology requires access to large datasets which can be cumbersome to curate and unwieldy to make use of. In this paper we present a suite of Python software packages which we collectively refer to as PERC (profet, EMPIARreader and CAKED). These are designed to reduce the burden which data curation places upon structural biology research. The protein structure fetcher (profet) package allows users to conveniently download and cleave sequences or structures from the Protein Data Bank or Alphafold databases. EMPIARreader allows lazy loading of Electron Microscopy Public Image Archive datasets in a machine-learning compatible structure. The Class Aggregator for Key Electron-microscopy Data (CAKED) package is designed to seamlessly facilitate the training of machine learning models on electron microscopy data, including electron-cryo-microscopy-specific data augmentation and labelling. These packages may be utilised independently or as building blocks in workflows. All are available in open source repositories and designed to be easily extensible to facilitate more advanced workflows if required.

1. **Introduction**

Cryogenic-sample Electron Microscopy (cryoEM) is an imaging technique used to obtain the structure of biomolecular objects at near-atomic resolution scales experimentally via transmission electron microscopy (TEM) of cryogenically frozen samples. Due to advancements in hardware and software since over the last decade, the resolution achievable via cryoEM reconstruction approaches that possible through x-ray crystallography [Cheng, 2015], with cryoEM being of particular use for determining the structure of macromolecules not amenable to other experimental methods such as x-ray crystallography or nuclear magnetic resonance (NMR) spectroscopy [Nogales, 2016]. The images which make up cryoEM datasets commonly have a very low signal to noise ratio (SNR) in order to minimise radiation damage. Consequently, the structures of macromolecules are typically obtained by averaging thousands of particle images which results in large volumes of data per structure.

The development of novel techniques in structural biology depends on the ability to access and utilise data and this can include synthetic data in some cases [Joosten et al., 2024, Jeon et al., 2024]. Experimentally reconstructed or predicted atomic models provide a vital tool for accelerating scientific software development in structural biology. Since its founding in 1971, >200,000 experimentally determined macromolecular structures have been deposited in the Protein Data Bank (PDB) archive [Berman et al., 2000]. The availability of this wealth of experimental data has been pivotal in the development of new software in the field, e.g. AlphaFold2 [Jumper et al., 2021].

Numerous approaches have used cryoEM datasets for the purpose of training deep learning models. Notable examples have been applied to the cryoEM image processing and reconstruction pipeline for particle picking [Bepler et al., 2019, Wagner et al., 2019], denoising [Bepler et al., 2020, Buchholz et al., 2019a, Buchholz et al., 2019b], 3D classification and dynamics [Zhong et al., 2021, Punjabi and Fleet, 2023, Schwab et al., 2024] and model building [Jamali et al., 2023, Si et al., 2020] among many more examples

[Chung et al., 2022]. The Electron Microscopy Public Image Archive (EMPIAR) [Iudin et al., 2022] is a public resource for the raw image data collected by cryoEM experiments and facilitates free access to this data, allowing it to be used for methods development and validation. Many of the resulting algorithms have been widely adopted as they enable quicker processing and/or improved interpretation of the data. There are also packages which allow for the simulation of synthetic micrograph datasets, such as Parakeet [Parkhurst et al., 2021] and MULTEM [Lobato and Dyck, 2015] which can be used to optimise data acquisition strategies [Parkhurst et al., 2024] and provide ground truth information to allow a greater understanding of the functioning of data processing algorithms [Joosten et al., 2024].

Being able to easily access these open repositories of experimental and simulated data is crucial for accelerating scientific software development in structural biology. However, in practice, doing this can be challenging, particularly for non-domain specialists or those new to the field. Each atomic model database has their own manual download system or individual Python package. On the other hand, deep learning-based approaches require large amounts of data to train the algorithms. For example, EMPIAR datasets can have hundreds of files and sizes on the order of terabytes or hundreds of gigabytes, meaning downloading and managing these datasets can become a barrier to the development of deep-learning methods. Additionally, the currently recommended tools to download data from EMPIAR either use proprietary software, require a user account or necessitate a web browser. As a result, data handling and management can become a largely manual task, limiting the ease of algorithm development, particularly for researchers new to the field.

In this paper, we present a suite of software packages which we have developed in order to address these issues: profet, EMPIARreader and CAKED. These packages may be used independently or utilised to build workflows, such as shown in Figure 1. With profet, users can conveniently download individual sequences directly using Python by simply specifying their Uniprot ID [The UniProt Consortium, 2022]. Users can specify which database they would like to use by default and, if the structure is available from that source, it will be downloaded. If the structure is not available from that source, profet will seek to download it from an alternative database (i.e., if the requested source is AlphaFold and the structure is not available, it then retrieves it from PDB). EMPIARreader is an open source tool which provides a Python library to allow lazy loading of EMPIAR datasets into a machine learning-compatible format. EMPIARreader additionally provides a simple, lightweight command line interface (CLI) which allows users to search and download EMPIAR entries using glob patterns or regular expressions and then download files via FTP or HTTP(S). With all the data accessed and downloaded, it is still necessary to load it into a machine learning readable format. For PyTorch, we have developed the Class Aggregator for Key Electron-microscopy Data (CAKED) software package. CAKED loads the images (both for single particle images or movies, and tomographic tilt series) from a local source. After loading and augmentation, the data is stored in a PyTorch DataLoader class ready to be used for training/classification. CAKED is designed to allow easy incorporation of different databases and is designed to facilitate seamless loading directly from the online data sources without local storage. This allows for seamless workflows (see Figure 1) where ML practitioners can access and utilise the data, whether from EMPIAR, Parakeet or elsewhere for training or deploying ML models. The software packages are open source and available to download from their GitHub repositories.

## 2. Software Packages

While the software packages can be used in the workflow exemplified in Figure 1, each package can be used individually for their defined purposes.

### 2.1. profet

The profet library provides a convenient unified interface to retrieve structures of biological macromolecules from either the PDB or AlphaFold database, simply by specifying the Uniprot ID (Figure 2). When a structure file is downloaded, it is cached to a local directory; if the same structure is requested again, either during the same session or a later session, then the cached structure file will be used preventing redundant downloads. Various potential applications require the ability to download many structures on demand including: protein matching algorithms for visual proteomics [Wagner et al., 2019, Mirecka et al., 2022], large scale models in molecular dynamics simulations [McGuffee and Elcock, 2010, Stevens et al., 2023], and electron microscopy simulations [Parkhurst et al., 2021]. In addition to a straightforward Python API, profet provides a simple command line interface, enabling the user to utilise profet either as part of a script or as a standalone program.

Commonly, structures are downloaded directly from the respective portal interfaces. For batch downloads, it is necessary to have scripting skills [RCSB PDB, 2024]. However, with computational power increasing and facilitating the use of large training sets, it is becoming increasingly important to automate the pipeline of PDB structure access from both experimental and simulated sources, which profet provides a portal to. As an added feature, profet provides the option to cleave the signal peptides from the retrieved structures (e.g. commonly found in AlphaFold structures). It also offers the deletion of hydrogens, water, or hetero atoms. Furthermore, profet is scalable and has the ability to add other databases as search options, such as CATH [Sillitoe et al., 2020] for example, by providing a template for accessing database APIs.

## 2.2. EMPIARreader

Raw cryoEM image datasets can be deposited into the online public image archive, EMPIAR [Iudin et al., 2022]. There is a loose schema to follow, but generally each deposited dataset is structured according to the needs or preferences of the depositing user with no particular directory tree enforced. With over 2040 entries and > 4.3PB of data hosted (as of October 2024), EMPIAR has become an important resource for the structural biology community, amassing over 700 citations in published works.

In addition to conventional reuse, such as to compare new software to popular entries, datasets from EMPIAR have been used extensively for training and validating cryoEM related deep learning algorithms, particularly for those which rely on raw image data. To make optimal use of the archive it is essential that the datasets are easily obtainable, and their size does not hinder accessibility or algorithm performance. The current recommended methods to download data from EMPIAR are via:

1. the IBM Aspera Connect web interface [IBM, 2023]

2. the IBM Aspera command line interface [IBM, 2020]

3. Globus [Foster, 2011, Allen et al., 2012]

4. HTTP(S) or FTP from the entry web page using an internet browser [Iudin et al., 2022]

These methods all require that data is downloaded and persisted before use and offer limited configurability and automation in data selection and access. To address this and to provide a way to integrate EMPIAR data into machine learning codebases, we have developed EMPIARreader, an open source tool which provides a Python library to allow lazy loading of EMPIAR datasets into a machine learning-compatible format (Figure 3). It parses EMPIAR metadata, uses the mrcfile library [Burnley et al., 2017] to interpret MRC files, supports common image file formats and uses the starfile library [Burt, 2020] to interpret STAR files. EMPIARreader additionally provides a simple, lightweight command line interface (CLI) which allows users to search and download EMPIAR entries using glob patterns or regular expressions and then download files via FTP or HTTP(S).

Therefore, in contrast to other methods, EMPIARreader allows data and metadata to be downloaded in a dynamic manner through lazy loading whilst also providing a simple interface for downloading EMPIAR files persistently to disk if required. EMPIARreader allows the granularity of downloads to be configured from an entire EMPIAR entry down to individual files. This makes EMPIARreader flexible enough to handle tasks from downloading a single file to downloading custom subsets of data from different EMPIAR entries. In principle, EMPIARreader allows any user to make use of the entire data archive without utilising local disk storage resources. It is envisioned that this utility will be particularly useful for the training of ML models and allow improved algorithmic performance by allowing fast and lightweight access to diverse training data. To see examples of the EMPIARreader API and CLI please refer to section 3.2 or the Jupyter Notebook (https://github.com/ccpem/empiarreader/blob/main/examples/run_empiarreader.ipynb) accessible on the EMPIARreader github page. EMPIARreader documentation can otherwise be found here (https://empiarreader.readthedocs.io/en/latest/).

### 2.3. CAKED

CAKED is a software package that decouples data loading and processing from the analysis (Figure 4). The package extends standard PyTorch [Ansel et al., 2024] Dataset and DataLoader primitives, and is itself extendable for more specific datasets and applications. After downloading cryoEM data, either to a local storage or in cache, it is necessary to represent it in a format suitable for machine learning. This includes both images files (which include their own metadata) and metadata files (such as picked particle metadata files).

CAKED is capable of processing both 2D and 3D image data in both MRC [Cheng et al., 2015] and Numpy [Harris et al., 2020] file formats. It assumes a file naming convention containing a class, followed by an underscore, followed by any other useful suffix information. As an extension of PyTorch Dataset class, it makes use of various existing PyTorch transforms as well as implementing a number of its own cryoEM/-ET specific preprocessing functions, including rescales, smoothing/blurring, and various normalisation methods. Furthermore, it is compatible with any other existing transforms available in torchvision [Marcel and Rodriguez, 2010]. The data is then wrapped in a PyTorch DataLoader class producing an iterable, along with labels and associated information related to each data point (e.g. filename suffix). The data is also automatically split into training and validation sets, as per the requested percentage split.

As the package is built in order to be scaled for large datasets, CAKED stores the paths instead of the data itself - this helps with the deployment of large datasets as each image is only accessed whenever the `get_item` function is called. For larger files, support for memmap objects will be added in future. While currently the methods implemented consider only the local storage, a template for other sources (for example, from cache) and the incorporation of different databases is available. Furthermore, this template could be extended for on-the-fly loading directly from the online databases bypassing the local storage in the future.

## 3. Applications

### 3.1. profet applications

#### 3.1.1. Installation

Install profet using pip:

```
pip install profet
```

To install the development version, which contains the latest features and fixes, installation can be directly done from GitHub using:

```
pip install git+https://github.com/ccpem/profet.git
```

To test the installation, you need to have pytest and pytest-cov packages installed which can be done as follows.

```
pip install pytest pytest-cov
```

Then navigate to the root directory of the package and run pytest. This code has been designed and tested for Python 3.

#### 3.1.2. Downloading PDB or Alphafold entries from Python

The profet library has a high-level Python API that can be used to download entries from both the PDB and AlphaFold through a single unified object oriented interface. An example of how to access this functionality through the main protein fetcher class, profet.Fetcher, is shown in the following code snippet.

```
from profet import Fetcher
# Initialise the fetcher and the cache directory
fetcher = Fetcher ( " pdb " , save_directory = " ~/. pdb / " )
# Get the filename and filedata
filename , filedata = fetcher . get_file ( " 4 v1w " , filetype = " cif " ,
filesave = True )
# Print the filename and first few lines of the file
print ( " Filename for 4 v1w = % s " % filename )
print ( " File contents : " )
print ( filedata [0:207]. decode ( " ascii " ))
```

```
# Get the search history
history_dictionary = fetcher . search_history ()
```

This results in the following output:

```
Filename for 4v1w = 4v1w.cif
File contents:
data_4V1W
#
_entry.id 4V1W
#
_audit_conform.dict_name mmcif_pdbx.dic
_audit_conform.dict_version 5.283
_audit_conform.dict_location
http://mmcif.pdb.org/dictionaries/ascii/mmcif_pdbx.dic
#
```

During initialisation, the default database can be specified in the constructor of the profet. Fetcher class.This should be a string containing either "pdb" for the PDB database or

"alphafold" for the AlphaFold database. In the example, the PDB database was specified. The directory to use for caching structure files can also be specified at this point by setting the `save_directory` keyword in the constructor. After initialisation, the fetcher can then be used to download structures from the specified database. This can be done by using the `profet.Fetcher.get_file` method and by specifying the id of the protein of interest. In the example, an apoferritin model ("4v1w") is downloaded, using the PDB id.

The universal identification value that works across platforms is the Uniprot ID. This is due to alphafold categorising structures only by their unique UniprotID, while PDB has a corresponding ID, as the same molecule can have different experimental entries. If the entry does not exist in any of the databases, then an exception is raised.

When downloading the protein structure, the `filetype` keyword can also be specified to choose between "cif" or "pdb" file if available. If the requested file type is not available then profet will attempt to download whichever file type is available. For example, if "cif" is requested but only "pdb" is available, the pdb file will be downloaded. If no file type is specified then the cif file will be download if present, otherwise the pdf file will be downloaded. Additionally, the `filesave` boolean flag can be used to specify whether or not the protein structure should be saved automatically to disk. By default this is set to False, in which case the returned filename is None and only the `filedata` is returned as a binary string. If `filesave` is `True`, then the file is saved into the current working directory. Finally, the `profet.Fetcher.search_history` function can be used to access the list of previously searched structures. The command will show a dictionary of the IDs searched by the fetcher and the databases where they are available as follows.

```
{'7U6Q': ['pdb'], 'F4HvG8': ['alphafold'], 'A0A023FDY8': ['pdb',
'alphafold']}
```

The functionality can be tested using the run `profet.ipynb` Jupyter notebook, available in the package repository.

### 3.1.3. Remove signal peptides and other functionalities

Once a structure is downloaded using `get_file`, the function `remove` from the `Fetcher` class lets you make a selection of things to remove from the structure file, often useful for simulation work. The option signal peptides compares the sequence of the structure to the

UniProt database for any signal peptides included in the structure. It then automatically deletes the signal peptides from the structure. Similarly, hydrogen atoms, water molecules, or hetero atoms can all be selected to be deleted from the structure. If no output name is given, the processed structure is saved as a separate file, with the options added to the filename (i.e. the deleted residue positions, or "nohydrogens").

```
import profet as pf
fetcher = pf . Fetcher ()
fetcher . set_directory ( " / path / to / directory / folder " )
fetcher . get_file (
uniprot_id = " P45523 " ,
filetype = " pdb " ,
filesave = True ,
db = " pdb "
)
fetcher . remove (
uniprot_id = " P45523 " ,
signal_peptides = True ,
hydrogens = True ,
water = True ,
hetatoms = True ,
output_filename = None
)
```

This will save p45523 1q6u.pdb and p45523 1q6u nosignal1to25 nohydrogens nowater nohetatm.pdb to the specified directory.

### 3.1.4. Downloading PDB or Alphafold entries from the command line

The profet library also has a command line interface that mirrors the Python API and which can be used to download entries from both the PDB and AlphaFold. An example of how to use the profet command line program is shown in the following code snippet.

```
profet 4v1w \
--filetype=pdb \
```

```
--main_db=pdb \
--save_directory="~/.pdb"
```

In this example, the entry "4V1W" is to be downloaded from the PDB database as a .pdb file. The file will be cached in the " /.pdb" directory for future use.

### 3.2. **EMPIARreader applications**

#### 3.2.1. **Installation**

EMPIARReader can be installed as a pypi package using Python ≥ 3.8 via:

```
pip install empiarreader
```

Otherwise, installation can be done with:

```
pip install git+https://github.com/ccpem/empiarreader.git
```

#### 3.2.2. **Using the EMPIARreader Python interface**

For this example, we open the EMPIAR entry 10943 and load an image dataset from its available directories.

```
from empiarreader import EmpiarSource , EmpiarCatalog
test_entry = 10943
# Reading XML into catalog
test_catalog = EmpiarCatalog ( test_entry )
# Retrieving catalog dataset from default directory
test_catalog_dir = list ( test_catalog . keys ())[0]
dataset_from_catalog = test_catalog [ test_catalog_dir ]
# Retrieving catalog dataset from specified directory
ds = EmpiarSource (
test_entry ,
directory = " data / MotionCorr / job003 / Tiff / EER / Images
- Disc1 / GridSquare_11149061 / Data " ,
filename = " .* EER \\. mrc " ,
regexp = True ,
)
```

```
# Read image
part = ds . read_partition (10)
```

Every EMPIAR entry has an associated XML file which contains the default order of the directory. This information can be accessed by loading the entry into an EmpiarCatalog. To get the dataset from the catalog, one would need to specify which directory to load. In the case above, there is only one so we choose the key in the position 0. However, the intended target is not always the directory present in the XML. We can further specify the directory to which directory we would like to get the images from. EMPIARreader can load the dataset from an EmpiarSource, using the EMPIAR entry number and the directory of the images. In the above case, we also specify that we want the MRC files from the specified directory. The dataset is loaded lazily using Dask [Rocklin, 2015], so the images are loaded one at a time when "read partition" is called. To choose an image, one can just pick the corresponding partition - for the example, partition 10 for the 11th image. This example can be visualised in the Jupyter Notebook provided in the EMPIARreader repository.

### 3.2.3. Using the EMPIARreader command line interface

You can use the EMPIARreader CLI to search the EMPIAR archive one directory at a time to find what you are looking for before then downloading those files to disk. First, you will need to choose an EMPIAR entry – in this example EMPIAR entry 10934 is used. Here we use a glob wildcard (–select "*") to list every subdirectory and file in a readable format:

```
empiarreader search --entry 10934 --select "*" --verbose
```

which returns:

```
Matching path #0:
https://ftp.ebi.ac.uk/empiar/world_availability/10934//10934.xml
Matching path #1:
https://ftp.ebi.ac.uk/empiar/world_availability/10934//data/
Subdirectories are:
https://ftp.ebi.ac.uk/empiar/world_availability/10934
Subdirectories are:
https://ftp.ebi.ac.uk/empiar/world_availability/10934//data
```

We've found the XML containing the metadata for the entry and a subdirectory called 'data'. To look inside you can add the '–dir' argument and repeat recursively until you find the directory you are interested in:

```
empiarreader search --entry 10934 --select "*" --dir "data" --verbose
```

Once you have found one or more files which you want to download from a directory in the EMPIAR archive you can create a list of URLs using the '–save search' argument:

```
empiarreader search --entry 10934 --dir \
"data/CL44-1_20201106_111915/Images-Disc1/GridSquare_6089277/Data" \
--select "*gain.tiff.bz2" --save_search saved_search.txt
```

Using the workflow described above, a user can quickly search and identify datasets that fulfill their criteria. These can then be downloaded using the download utility of the CLI. A user simply needs to specify the file list and a directory to download the files into:

```
empiarreader download --download saved_search.txt --save_dir new_dir --verbose
```

### 3.3. CAKED applications

#### 3.3.1. Installation

CAKED can be installed by cloning the latest version on GitHub:

```
git clone https://github.com/ccpem/caked
cd caked
python -m pip install .
```

This version has been tested with Python 3.

#### 3.3.2. Local loading of data

In this section, we will demonstrate the CAKED application considering the local storage of a dataset. Each image is stored with the respective class as the prefix of the name file. The first step is to create a "DiskDataLoader" object. Any required arguments are passed to the object during instantiation and are all optional: the pipeline from which to load (during the time of writing, only "disk" is available), the list of classes in the dataset, size of the dataset, whether the dataloader is used for training (default is True), the list of transforms to apply to the data and, finally, whether to save to disk the data after processing or no.

```
caked_loader = DiskDataLoader (
pipeline = DISK_PIPELINE ,
classes = LIST_OF_CLASSES ,
```

```
dataset_size = DATASET_SIZE ,
training = True ,
transformations = LIST_OF_TRANSFORMS ,
)
caked_loader . load ( datapath = PATH_TO_DATA , datatype = DATATYPE )
image , label = next ( iter ( caked_loader . dataset ))
```

After the loader has been instantiated, the data can be loaded using the load function. In this case, the path to the data must be provided, as well as the datatype (default is mrc). When the list of transformations in the loader is not empty, the process function is called in order to apply the transformations to the data. Finally, images can be accessed with iterator functions such as `next()` and `iter()`.

## 4. Discussion

The democratisation of cryoEM data into a machine learning-compatible pipeline is an important step in the automation of processes to further analyse the data. In this paper, we presented three packages that can be used to deal with manual bottlenecks in the data curation process. With profet, users can now download protein structures from both experimental or simulated data and feed them straight into their simulation or modelling pipelines. EMPIARreader enables lazyloading of large datasets downloaded from the online archive onto a Python format such as a numpy array, or direct download to local storage. Finally, CAKED can load different types of images into a PyTorch DataLoader which can be directly used with PyTorch-compatible models. Importantly, all three of these packages can be used as part of the data flow within a single software pipeline which needs, for example, both experimental and simulated data injected into a model (e.g., for generative model studies). Furthermore, the seamless integration of all three packages can be performed within a Python workflow. Further developments are also possible given the modularity of the tools: profet has a template to easily integrate other databases (for example, CathDB [Sillitoe et al., 2020]), EMPIARreader can be an off-the-shelf Python API tool for EMPIAR and CAKED will be integrated in the CCP-EM cryoEM ML toolbox. All the source code is open source (EMPIARreader - BSD 3-Clause "New" or "Revised" License, profet and CAKED – MIT License) and development is performed collaboratively, therefore it is possible for users of the packages to add other sources of data, and external contributions to the projects are welcome (see the contribution pages in each repository: profet, EMPIARreader, CAKED). It is our hope that these tools will simplify the process of accessing data and training ML models or performing any other computational analysis and reduce the setup cost for researchers in the field.

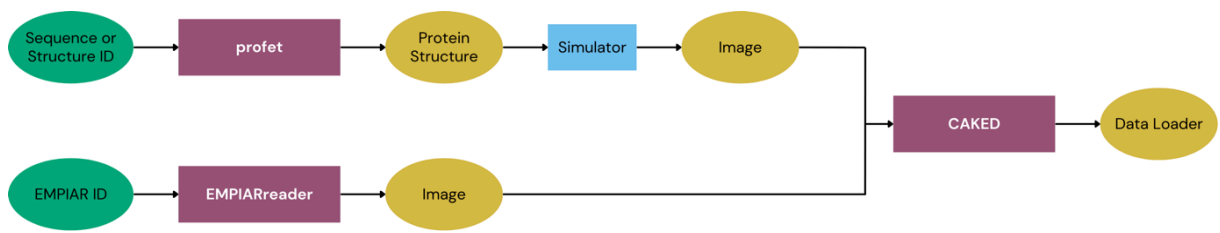

**Figure 1**   Flowchart of the overall pipeline utilising profet, EMPIARreader, and CAKED. Data can be retrieved from the preferred source with profet (and then input into a simulator to achieve a synthetic image populated with the structure); EMPIARreader allows lazyloading of the data from the specified directory, then further processing can be done with CAKED. Green corresponds to the inputs, purple to the implemented packages, blue to the outside sources, and yellow to the output.

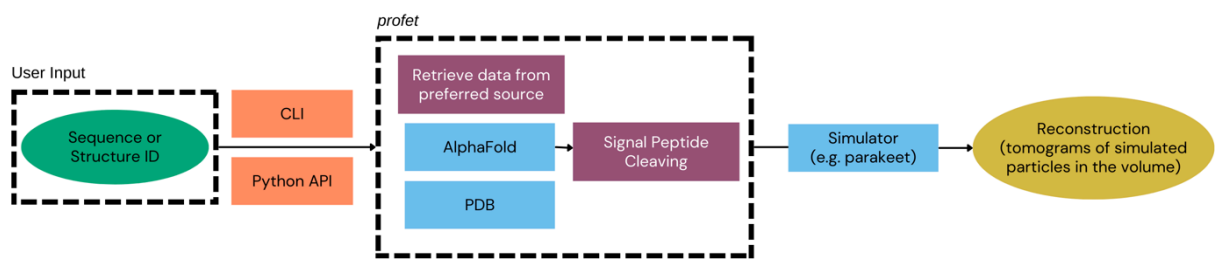

**Figure 2**   Flowchart of the profet pipeline. Green corresponds to the inputs, orange to the access type, purple to the implemented packages, blue to the outside sources, and yellow to the output. The user can input a protein structure or protein sequence ID and profet will provide the matching output data from the selected source after optional signal peptide cleaving for further processing or utilisation by the user.

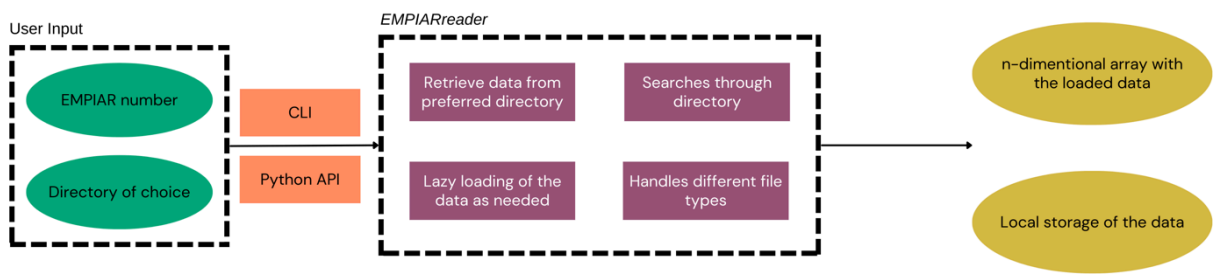

**Figure 3**    Flowchart of the EMPIARreader pipeline. Green corresponds to the inputs, orange to the access type, purple to the implemented packages, and yellow to the output. EMPIARreader allows data and metadata to be downloaded in a dynamic manner through lazy loading while also providing a simple interface for downloading EMPIAR files persistently to disk if required.

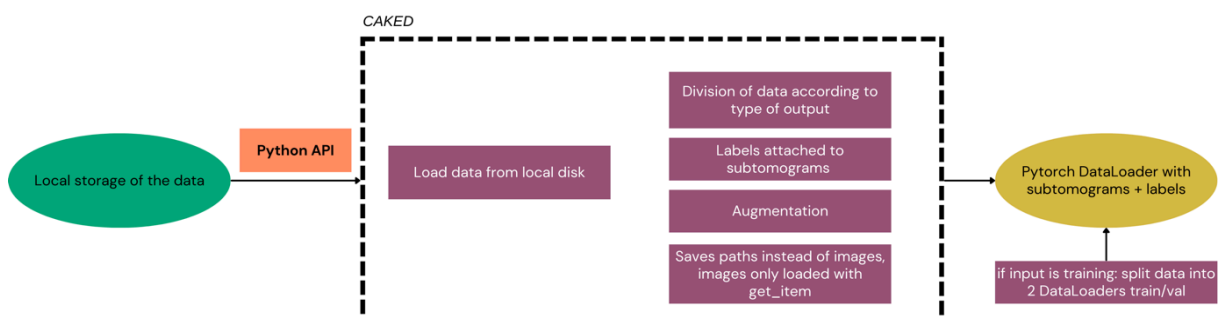

**Figure 4**    Flowchart of the CAKED pipeline. Green corresponds to the inputs, orange to the access type, purple to the implemented packages, and yellow to the output. CAKED is a software package that decouples data loading and processing from the analysis, and represents the data in a format suitable for machine learning. CAKED extends standard PyTorch Dataset and DataLoader primitives, and is itself extendable for more specific datasets and applications.

**Acknowledgements**    The Rosalind Franklin Institute is an EPSRC core funded Institute. Nikolai Juraschko is funded by EPSRC grant number EP/V521899/1 and this work was supported by The Alan Turing Institute's Enrichment Scheme. This work was supported by Wave 1 of The UKRI Strategic Priorities Fund under the EPSRC Grant EP/W006022/1, particularly the "AI for Science" theme within that grant & The Alan Turing Institute. Joel Greer, Jola Mirecka and Tom Burnley would also like to thank MRC Partnership Grant MR/V000403/1 and acknowledge support from the Ada Lovelace Centre for this work.

**Conflicts of interest** The authors declare that there are no conflicts of interest.